\documentclass{acmtog} 

\acmVolume{}
\acmNumber{}
\acmYear{}
\acmMonth{}
\acmArticleNum{}
\doi{}

\setcopyright{none}

\begin{document}

\markboth{T. Trein}{Siamese Networks for Cat Re-Identification: Exploring Neural Models for Cat Instance Recognition}

\title{Siamese Networks for Cat Re-Identification: Exploring Neural Models for Cat Instance Recognition} 

\author{TOBIAS TREIN
\affil{Pontifícia Universidade Católica do Rio Grande do Sul}
\and
LUAN FONSECA GARCIA
\affil{Pontifícia Universidade Católica do Rio Grande do Sul}
}

%
%


%
%



\acmformat{Tobias Trein. 2024. Siamese Networks for Cat Re-Identification: Exploring Neural Models for Cat Instance Recognition.} 

\maketitle

\begin{abstract}
Street cats in urban areas often rely on human intervention for survival, leading to challenges in population control and welfare management. In April 2023, Hello Inc., a Chinese urban mobility company, launched the Hello Street Cat initiative to address these issues. The project deployed over 21,000 smart feeding stations across 14 cities in China, integrating live-streaming cameras and treat dispensers activated through user donations. It also promotes the Trap-Neuter-Return (TNR) method, supported by a community-driven platform, HelloStreetCatWiki, where volunteers catalog and identify cats. However, manual identification is inefficient and unsustainable, creating a need for automated solutions.
This study explores Deep Learning-based models for re-identifying street cats in the Hello Street Cat initiative. A dataset of 2,796 images of 69 cats was used to train Siamese Networks with EfficientNetB0, MobileNet and VGG16 as base models, evaluated under contrastive and triplet loss functions. VGG16 paired with contrastive loss emerged as the most effective configuration, achieving up to 97\% accuracy and an F1 score of 0.9344 during testing. The approach leverages image augmentation and dataset refinement to overcome challenges posed by limited data and diverse visual variations.
These findings underscore the potential of automated cat re-identification to streamline population monitoring and welfare efforts. By reducing reliance on manual processes, the method offers a scalable and reliable solution for community-driven initiatives. Future research will focus on expanding datasets and developing real-time implementations to enhance practicality in large-scale deployments.
\end{abstract}

\section{introduction}
According to \cite{Hello2023}, \begin{CJK*}{UTF8}{gbsn} "哈啰街猫", \end{CJK*} or "Hello Street Cat," is a philanthropic initiative developed by Hello Inc., a Chinese urban mobility company, launched in April 2023. The initiative involves the installation of over 21,000 smart feeding stations for street cats across 14 cities in China. These feeding stations are equipped with cameras that transmit real-time video through the app, as well as dispensing treats every time a donation is made by users \cite{Skyrina2024}. The main goal of the app is to promote Trap-Neuter-Return(TNR), a street cat population control technique involving the capture, neutering, and release of these cats \cite{schmidt2009evaluation}. All funds raised from donations are used for maintaining the app and implementing TNR in areas where the feeding stations have been installed, with over 25,000 cats having already benefited from the initiative.

The app's engagement by Chinese users has led to the creation of virtual communities in China to share comedic videos featuring the reactions of cats using the feeding stations. The app's popularity outside of China began in October 2023, when memes created by the Chinese community were shared by an American TikTok account. One of the English-speaking communities, the HelloStreetCatWiki, created in February 2024, maintains an encyclopedia that names, describes, and logs the appearances of cats at the most popular feeding stations\cite{Skyrina2024}. 

Within the wiki, each cat has its own page containing general information such as name, photos, and a log of their appearances, which is collaboratively updated by users. This identification work is done manually using an identification list that categorizes animals by color \cite{thefloppypig2024}. Manual identification is prone to human errors, such as confusion between similar-looking cats or outdated appearance logs. Keeping this log updated requires constant effort from contributors, which can be difficult to maintain over time. Automating this process using deep learning models for instance recognition can solve these problems by providing faster, more accurate, and scalable identification, allowing contributors to focus on other activities that add greater value to the community.

For this reason, this study proposes an investigation into deep learning approaches to address the problem of animal re-identification. Specifically, we aim to evaluate the performance of different neural network models within the Siamese network framework, which is commonly used for instance recognition. The dataset used in this study comprises images of street cats collected from the Hello Street Cat initiative. Additionally, we assess these models under different loss functions, and evaluate the impact of the type of input images used, specifically the collected top and front perspectives. Techniques like image augmentation are also explored to address dataset limitations and enhance model performance.

The remainder of this paper is organized as follows: Section \ref{sec:theorectical} outlines the Theoretical Background, detailing the core concepts and techniques employed in this study. Section \ref{sec:related} provides an overview of the Related Work, summarizing key studies and advancements in individual animal recognition. Section \ref{sec:dataset} introduces the Dataset, describing its creation and characteristics. Section \ref{sec:experiments} explains the Experiments, including the methodologies and configurations used. Section \ref{sec:results} presents the Results \& Discussion, analyzing the outcomes and their implications. Finally, Section \ref{sec:future} outlines potential Future Works to expand and enhance this research area.

\section{Theorectical Background}
\label{sec:theorectical}
In this section, we present the foundational concepts and techniques that underpin the development of our proposed approach. These concepts provide the necessary theoretical framework for understanding the methods employed in this work. On the following subsections \ref{subsec:vgg}, \ref{subsec:efficient}, and \ref{subsec:mobile}, we detail the specific models employed, highlighting their architectures and explaining their relevance to the task at hand.

\subsection{Siamese Networks and CNNs}
One of the most relevant approaches in animal identification tasks is the use of Siamese networks. According to \citet{Chicco2021}, "it consists of two identical artificial neural networks that shares the same weights, each capable of learning the hidden representation of an input vector. The two neural networks are both feedforward perceptrons, and employ error back-propagation during training; they work parallelly in tandem and compare their outputs at the end, usually through a cosine distance. The output generated by a siamese neural network execution can be considered the semantic similarity between the projected representation of the two input vectors."

Siamese networks have been successfully applied to a variety of re-identification tasks. For instance, they have been employed for cat recognition \cite{li2022cat}, human identification \cite{PEI2023109148}, and goat recognition \cite{su2022intelligent}, showcasing their adaptability across different species. The ability of Siamese networks to compare pairs of inputs and emphasize distinguishing features makes them particularly well-suited for identifying individuals in datasets with high visual variability.

In addition to Siamese networks, Convolutional Neural Networks (CNNs) remain a cornerstone of computer vision tasks, including animal identification. CNNs utilize convolutional layers, where a kernel or filter, typically a small matrix of weights (e.g., 3x3), slides over the input image. This process detects local patterns, such as edges or textures, by performing convolution operations and generating feature maps. These maps capture spatial hierarchies of features, enabling CNNs to learn complex patterns from raw image data.

CNNs have been widely applied to animal identification tasks. For example, traditional CNN architectures have been used for chimpanzee recognition \cite{schofield2019chimpanzee} and cat identification \cite{cho2023multi}. Their ability to automatically extract hierarchical features from images makes them highly efficient for tasks like object detection, image segmentation, and individual recognition.

By combining the strengths of both Siamese networks and CNNs, researchers have developed robust frameworks for identifying and re-identifying animals across various datasets and applications. These approaches continue to enhance the precision and scalability of animal identification in the field of computer vision.

\subsubsection{VGG16}
\label{subsec:vgg}
Designed to improve image recognition tasks, VGG16 is a deep Convolutional Neural Network (CNN) architecture designed to improve performance by increasing depth while maintaining a consistent and straightforward structure \cite{simonyan2014very}. 

The model comprises 13 convolutional layers, 5 max-pooling layers, and 3 fully connected (dense) layers, resulting in 16 layers with learnable parameters, which is where the name "VGG16" originates. Each convolutional layer employs small 3x3 filters with a stride of 1, combined with "same" padding to preserve spatial dimensions, ensuring precise feature extraction. 

Additionally, the max-pooling layers use 2x2 filters with a stride of 2 to reduce spatial dimensions while retaining essential information. This systematic approach to layer design, focusing on small filter sizes and depth, was a significant factor in its success, demonstrating substantial improvements over earlier architectures.

\subsubsection{EfficientNet}
\label{subsec:efficient}
A groundbreaking approach to CNN design, EfficientNet is a convolutional neural network architecture designed to optimize scaling across depth, width, and resolution by using a compound coefficient \cite{tan2019efficientnet}.

Unlike traditional methods that scale these dimensions independently, EfficientNet employs a compound scaling strategy controlled by a coefficient \(\phi\). This coefficient determines how the depth, width, and resolution are adjusted, with each dimension scaled by \(\alpha^\phi\), \(\beta^\phi\), and \(\gamma^\phi\), respectively. 

These constants \(\alpha\), \(\beta\), and \(\gamma\) are determined through a grid search on a baseline model, ensuring efficient use of computational resources. For instance, doubling computational power (\(2^\phi\)) allows the network to scale proportionally across all dimensions, maintaining a balance between performance and efficiency. This unified scaling method enhances accuracy while reducing unnecessary complexity in the network design.

\subsubsection{MobileNet}
\label{subsec:mobile}
For resource-constrained devices, MobileNet is a new approach to convolutional neural networks focused on reducing model size and computational cost through depthwise separable convolutions. By separating the convolution process into two distinct operations—depthwise convolution and pointwise convolution—this model significantly reduces the number of parameters compared to traditional convolutions. 

Depthwise convolutions apply a separate filter for each input channel, while pointwise convolutions use 1x1 filters to combine the results from the depthwise convolutions. This separation minimizes computational cost and improves efficiency, making the model particularly suitable for mobile and edge devices with limited processing power. The approach allows for fast processing while maintaining a small model size, providing a powerful solution for real-time image classification tasks on mobile devices \cite{howard2017mobilenets}.

\subsection{Data Augmentation}
Data augmentation is a technique used to artificially increase the size of a dataset by generating additional training samples. This can be achieved through transformations applied directly in the data-space, such as rotations, flips, or scaling of images, or by creating synthetic samples in the feature-space. These methods enhance the diversity of training data, which helps machine learning models generalize better.

According to \citet{wong2016understanding}, "data augmentation acts as a regularizer in preventing overfitting in neural networks and improves performance in imbalanced class problems." Experiments on models like convolutional neural networks (CNNs), show that plausible transformations in data-space offer greater benefits compared to feature-space augmentation. This approach has been shown to improve the performance of classifiers, reducing testing error and overfitting, making it an essential technique for training robust machine learning models.

\subsection{Transfer Learning}
Transfer learning is a machine learning technique where a model developed for one task is reused as the starting point for a model on a second task. This approach is particularly useful when the second task has limited data, as it allows the model to leverage knowledge gained from a related domain. 

In transfer learning, the model is first pre-trained on a large dataset, such as ImageNet \cite{Deng2009}, which contains millions of images across thousands of categories. By learning general features in the pretraining phase, the model can then be fine-tuned on the smaller, task-specific dataset \cite{bengio2012deep}. 

This process significantly improves performance, especially in cases where labeled data is scarce, as the model is able to transfer the learned representations to new, but related tasks \cite{keras2023}. Transfer learning is widely used in computer vision and natural language processing, where large, well-labeled datasets are available for pre-training, and the learned knowledge can be applied to various downstream tasks.

\section{Related Work}
\label{sec:related}
The individual identification of animals through images has been a challenge faced by researchers in many fields, such as biology, ecology, and computer science. With advances in technology, deep learning techniques have emerged as promising tools for addressing this problem. Several re-identification techniques have been used for at least five years to achieve individual animal recognition, particularly in the agricultural sector. In \cite{ravoor2020deep}, the authors conducted a survey analyzing the main deep learning approaches for animal re-identification. They categorized these techniques into two distinct strategies: localized parts-based approaches and face and head-based approaches. Parts-based methods focus on distinctive features like fur patterns or body parts, while face and head-based approaches use facial landmarks and unique head characteristics for recognition. The choice of strategy depends on image quality and variability in the animals' appearances.

\subsection{Localized parts detection}
Localized parts-based methods focus on identifying distinctive characteristics from specific areas of an animal’s body, such as black and white patterns on the backs of cows, the unique shapes of dolphin fins, or the color patterns of minke whales. For instance, Phyo et al. \cite{phyo2018hybrid} utilized a 3D-DCNN model to analyze the back patterns of cows, while Bouma et al. \cite{bouma2018individual} relied on a ResNet-based model to differentiate dolphins based on their fins. Konovalov et al. \cite{konovalov2018individual} applied the VGG16 neural network to recognize minke whales by their distinctive color patterns. These methods effectively leverage advanced deep learning architectures to capture the unique traits that distinguish individuals within a species.

In terms of results, localized parts-based detection methods have demonstrated impressive performance across various species. For minke whales, the approach achieved an F1 score of 0.76. For cows, the method reported an overall accuracy of 96.3\%, while for dolphins, a top-5 accuracy of 93.6\% was achieved. These results highlight the effectiveness of localized parts-based methods in identifying and differentiating individuals based on distinctive physical traits.

\subsection{Face and head detection}
On the other hand, face and head-based approaches operate under the assumption that facial features alone are sufficient for individual identification. This is particularly relevant for species where unique and distinguishable traits, such as patterns, shapes, or textures, are predominantly concentrated in the facial region. For instance, \cite{li2018cow} applied a DnCNN model to identify cows based on facial structures, while \cite{he2019distinguishing} used a VGG16 model to achieve accurate identification of red pandas by analyzing their facial features. These studies highlight how distinct traits in the facial region can be effectively leveraged with advanced deep learning models to differentiate individuals.

In terms of performance, face and head-based detection methods have shown impressive results. For cows, the method achieved an accuracy of 95\% for the top-3 predictions. For red pandas, the technique reached a high accuracy of 98.3\% at rank 10, underscoring the power of facial recognition in identifying individuals based on their unique facial features.

\subsection{Cat re-identification}
Specific to cats, research on individual identification has shown promising results with different techniques and detection strategies. In \cite{li2022cat}, the authors focused on facial detection and utilized a VGG16-based model combined with Siamese networks, achieving an accuracy of 72.91\%. Fan et al. \cite{fan2021cat} employed face detection techniques alongside Mel-Frequency Cepstral Coefficients (MFCC) and Gaussian Mixture Models (GMM), reaching an accuracy of 83.3\%. More recently, Cho et al. \cite{cho2023multi} advanced the field by leveraging both face and body detection, using EfficientNetV2 as a feature extractor paired with a Support Vector Machine (SVM) classifier, achieving a remarkable accuracy of 94.53\%. These studies underscore the effectiveness of various detection and classification approaches for cat re-identification, highlighting advancements in feature extraction and detection strategies.

\begin{table}[ht]
\tbl{Performance metrics for animal instance recognition in related works. \label{tab:rel}}{
\centering
\vspace{0.3em} 
\begin{tabular}{ll}
\hline
\textbf{Localized parts detection}                      & \textbf{}        \\ \hline
\cite{phyo2018hybrid} cows             & 96.3\%  accuracy \\
\cite{bouma2018individual} dolphins    & 93.6\% accuracy  \\
\cite{konovalov2018individual}whales   & 0.76 F1 score    \\ \hline
\textbf{Face and head detection}                        & \textbf{}        \\ \hline
\cite{li2018cow} cows                  & 95\% accuracy    \\
\cite{he2019distinguishing} red pandas & 98.3\% accuracy  \\ 
\hline
\textbf{Cat Re-identification}                        & \textbf{}        \\ \hline
\cite{li2022cat}  & 72.91\% accuracy  \\
\cite{fan2021cat}  & 83.3\% accuracy  \\
\cite{cho2023multi} & 94.53\% accuracy  \\
\hline
\end{tabular}}
\end{table}

\section{Dataset}
\label{sec:dataset}

One of the contribution of our work is the HelloStreetCat Individuals dataset. Publicly available on Kaggle \footnote{https://www.kaggle.com/datasets/tobiastrein/heellostreetcat-individuals}, consists of 2,796 images of 69 cats, with an average of 41 images per cat. All images were captured exclusively at The Happy Canteen Feeder, the feeder with the highest amount of data and named cats. This dataset was created specifically for this study and shared on Kaggle, a Google-established platform for practicing machine learning concepts and publishing datasets \cite{Kaggle2024}.

The images are organized into folders named after the active cats registered on the StreetCatWiki. Each folder represents a specific cat and includes 2 subfolders for the two types of images collected: frontal-view images, stored in a subfolder named front/, and top-view images, stored in a subfolder named top/.

The images were taken from different angles and at various times of the day to ensure a broader and more diverse representation of each cat's appearance under different conditions. This approach was aimed at improving the robustness and variability of the dataset. 

To feed this dataset, we developed HelloStreetCat Live Scraper\footnote{https://github.com/TobiasTrein/hsc-live-scrapper}, a Python script to automate the process of capturing screenshots from the "HelloStreetCat" livestream on YouTube. Using YOLOv8-cls \cite{yolo2023}, an object detection model that simultaneously localizes and classifies multiple objects in an image model \cite{Jocher2023}, the script was designed to analyze the livestream in real-time and capture images only if a cat was detected. The approach integrates video capture functionality using YT-DLP media extractor library \cite{pukkandan2023} with real-time image processing with FFmpeg \cite{Bellard2000} to retrive the images, ensuring that only relevant moments, where the cat is visible, are recorded. After that, the images need to be cropped and divided into their respective folders.

To run the script, the user only needs to provide the URL of the livestream as a command-line argument. By leveraging YOLO, the script enables accurate real-time detection of cats, providing an efficient and hands-off way to generate data for our dataset.

\begin{figure}[ht]
\centerline{\includegraphics[width=.48\textwidth]{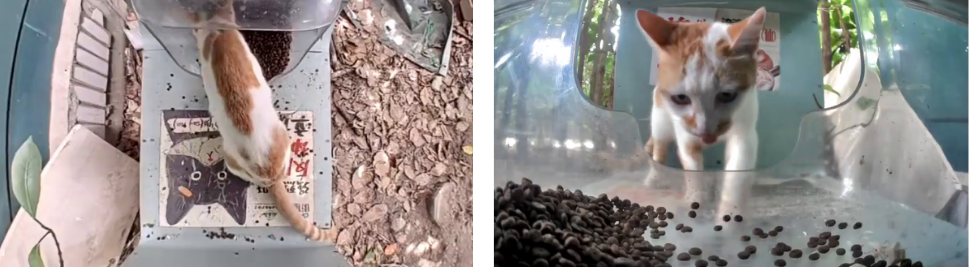}}
\caption{Example images of Mr. Egg used in the dataset. The left image shows a top view, while the right image shows a front view, illustrating the two perspectives captured for each individual cat.}
    \label{fig:mr-egg}
\end{figure}

\section{Experiments}
\label{sec:experiments}
The Experiments section details the practical steps taken to develop, train, and evaluate the Siamese Network model proposed in this study. This includes the configuration of the computational environment, the preparation of the dataset, and the implementation of the model.

\subsection{Experimental Setup}
For the execution of our model, a local environment was configured using Docker, an open platform for developing, shipping, and running applications. Docker allows applications to be isolated from their infrastructure through lightweight units called containers \cite{Docker2024}. Within this container, we installed the NVIDIA Container Toolkit. "The NVIDIA Container Toolkit enables users to build and run GPU-accelerated containers. This toolkit includes a container runtime library and utilities that automatically configure containers to leverage NVIDIA GPUs" \cite{Nvidia2024}. 

The local machine where this environment is configured is equipped with a GeForce RTX 4050 graphics card with 6 GB of memory, providing the necessary performance for executing the intended operations.

\begin{figure}[ht]
\centering\includegraphics[width=.45\textwidth]{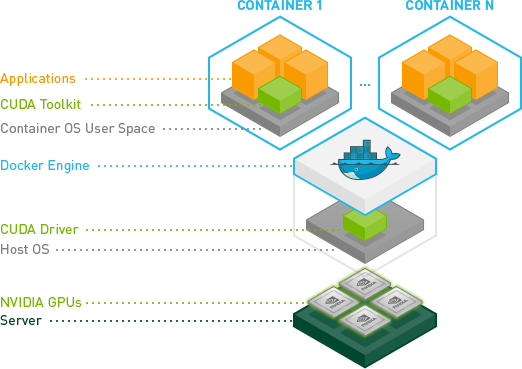}
\caption{\label{fig
}Graphical visualization of the environment}
\end{figure}

\subsection{Dataset Preparation}
\label{sec:dataset_prep}

In Section~\ref{sec:dataset}, we described the curation of a dataset tailored to our experiments. For the tests with the Siamese Network model, we utilized a subset of cat instances that had at least 40 images each. This selection criterion ensured the availability of sufficient data for effective model training and evaluation.

The dataset was specifically designed to facilitate flexibility in experimentation for front, top, and both image types. During model execution, this organization enables users to easily specify the type of images to use, allowing for comprehensive evaluation across varied perspectives and input configurations.

To ensure a balanced distribution of classes across all splits, the dataset was divided into training, validation, and test sets following an 80:10:10 ratio to ensure a balanced distribution of classes across splits. Initially, 20\% of the data was separated for testing using stratified sampling to preserve class proportions. Subsequently, the test set was evenly divided into validation and test subsets, maintaining the class distribution. This ensures a reliable performance evaluation while preventing data leakage between the training and evaluation phases.

\subsection{Implementation Overview}
For the development of a Siamese Network model, we used as the basis for the initial code an adaptation of the individual cat\_snn implementation \cite{Luo2021}, available on Kaggle. 
Building upon this implementation, we created a more flexible version that accepts multiple hyperparameters and executes all the combinations selected sequentially, significantly streamlining the experimentation process by enabling automated testing of various configurations without manual intervention. These hyperparameters are the Photo Type (front, top, all), Base Model (VGG16, MobileNet, EfficientNet), Loss Function (Contrastive, Triplet), Number of Epochs, Learning Rate, and Augmentation (none, flip, noise, rotate).

\subsubsection{Implementation Architecture}
The solution's architecture\footnote{https://github.com/TobiasTrein/hsc-reident/} is based on a Siamese Network, comprising two independent subnetworks that share the same weights and are responsible for generating embeddings from input images. These embeddings are compared through a Lambda layer which calculates the Euclidean distance, providing a similarity measure between the two input instances.

\begin{figure}[ht]
\centering\includegraphics[width=.46\textwidth]{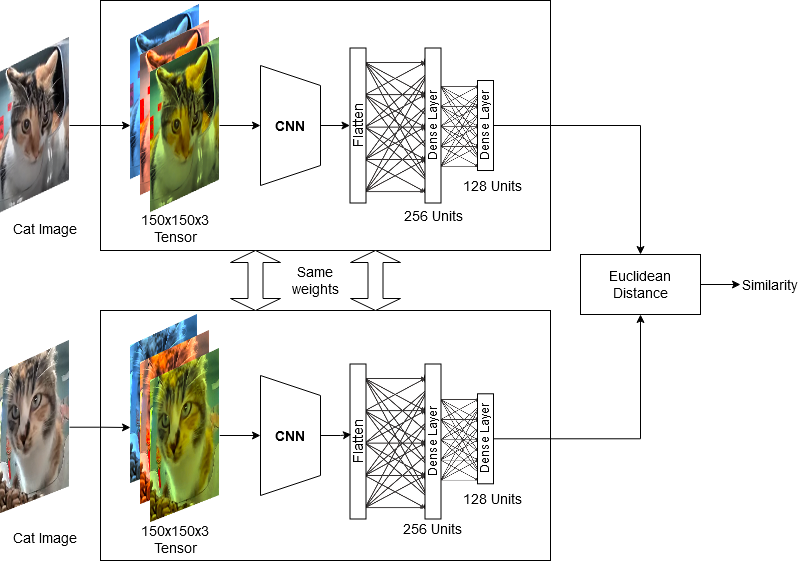}
\caption{\label{fig:snn} Simplified visualization of the Siamese Network model using Mr. Chess and Ms. Princess as example}
\end{figure}

Each independent network is composed of multiple layers, with the first layer responsible for data input. This initial layer consists of a three-dimensional tensor designed to receive images with dimensions of 150x150 pixels, divided into three channels corresponding to the RGB color system.

The choice of 150x150 pixel dimensions for the input balances performance and computational efficiency, ensuring that the model can process images effectively without being overwhelmed by the high number of elements present in larger dimensions. This size provides sufficient detail for feature extraction while optimizing resource usage during training and inference.

The second layer of each subnetwork includes an instance of a CNN as a feature extractor. It begins with a pre-trained CNN model implemented using the Keras library. Depending on the configuration, these CNNs can be VGG16, MobileNet, or EfficientNet. 

Each of these architectures includes several convolutional and pooling layers that extract hierarchical features from the input images. Their fully connected classification layers are removed to make the models suitable for embedding generation. This modification transforms these architectures from classification models into feature extractors tailored to our task. To further optimize performance, we adopt transfer learning, freezing the weights of the pre-trained layers to retain their previously learned feature extraction capabilities.

The custom components of the network architecture are added after the output of the base CNN. First, the three-dimensional tensor output from the CNN is processed through a Flatten layer, converting it into a one-dimensional vector. Next, a Dense layer with 256 units and ReLU activation is applied, enabling the network to learn complex, non-linear representations from the extracted features. Finally, the output layer, another Dense layer with 128 units, produces the embedding vector. This layer maps the learned features to a reduced-dimensional space that encodes the key information necessary for the similarity calculations.

The two subnetworks process their respective inputs independently, generating two embeddings. These embeddings are then fed into a Lambda layer, which computes the Euclidean distance between them. This distance serves as the output of the network and reflects the similarity between the two input images.

The Siamese Network architecture described is employed to recognize a specific instance of a cat by comparing a test image with a set of pre-existing reference images (anchors), composed by one image per instance. The test image is passed through the network alongside each anchor image. The network generates embeddings for both the test image and each anchor image, and the Euclidean distance between these embeddings is calculated. This distance serves as a measure of similarity, indicating how closely the test image matches each anchor.

If any of the distances are below a predefined threshold (i.e., 0.4), the model identifies the test image as belonging to one of the known cats, and the corresponding cat ID is found. If no match is found (i.e., if all distances are above the threshold), the test image is classified as an "unknown" cat. This approach leverages the Siamese Network's ability to generate meaningful embeddings, allowing for effective re-identification of individual cats based on similarity measures.

The described architecture provides flexibility, allowing us to explore and compare the performance of different CNN backbones (VGG16, MobileNet, and EfficientNet) while leveraging the consistent embedding generation and distance computation layers for similarity assessment. Figure \ref{fig:snn} illustrates this architecture..

\subsubsection{Cost Functions}
The cost function for training the Siamese network architecture can be selected from either the Contrastive Loss or Triplet Loss functions, depending on the specific requirements of the experiment. 

The Contrastive Loss function aims to minimize the distance between embeddings in the feature space if they represent the same cat. For instances from different cats, it enforces a minimum distance margin by penalizing the model when the embeddings of such instances are closer than the specified margin \cite{hadsell2006dimensionality}.

Alternatively, the Triplet Loss function operates by comparing triplets of instances: an anchor, a positive sample (same class as the anchor), and a negative sample (different class from the anchor). This loss encourages the model to bring the anchor closer to the positive sample in the feature space while pushing it farther from the negative sample, ensuring better separation and clustering of embeddings.

\subsubsection{Libraries and Tools}
To create the models, we utilized the following main libraries: TensorFlow v2.16.2 and its integrated Keras v3.4.1 library, which provide high-level APIs for building, training, and evaluating models \cite{Keras2023a}; Pandas v2.2.2 for tabular data manipulation and analysis \cite{Pandas2024}; Scikit-learn v1.5.0 for data preprocessing and dataset splitting \cite{scikitlearn2024}; OpenCV v4.10.0 for image manipulation and processing \cite{OpenCV2024}; and Matplotlib v3.9.0 for data visualization and result presentation \cite{matplot}.

\section{results \& discussion}
\label{sec:results}
The evaluation of our approach was carried out in multiple stages, beginning with initial experiments conducted during the early phases of dataset creation. At this stage, the dataset contained fewer cat instances meeting our criteria described in section \ref{sec:dataset_prep} of having at least 40 images per instance. 

\subsection{Initial Tests}
In the initial tests, we determined that 100 epochs were sufficient for training the model, as extending to 200 epochs did not lead to any performance improvement. These tests were conducted on a smaller subset of 10 instances, allowing for a preliminary evaluation of the model's performance. 

However, the results with EfficientNetB0 as shown on table ~\ref{tab:efi} were unsatisfactory, with the highest accuracy achieved being only 36\% for the top view using the contrastive loss function. Other configurations, such as the triplet loss function or the front view images, yielded even lower accuracies, indicating that EfficientNetB0 is not well-suited for our architecture.

\begin{table}[ht]
\tbl{Results of preliminary executions using EfficientNet with different loss functions and photo types.\label{tab:efi}}{
\centering
\vspace{0.3em} 
\begin{tabular}{llll}
\hline
\rowcolor[HTML]{FFFFFF} 
\multicolumn{1}{c}{\cellcolor[HTML]{FFFFFF}\textbf{Photo Type}} & \multicolumn{1}{c}{\cellcolor[HTML]{FFFFFF}\textbf{Base Model}} & \multicolumn{1}{c}{\cellcolor[HTML]{FFFFFF}\textbf{Loss Function}} & \multicolumn{1}{c}{\cellcolor[HTML]{FFFFFF}\textbf{Accuracy}} \\ \hline
\rowcolor[HTML]{FFFFFF} 
\textbf{top}                                                    & \textbf{efficientNetB0}                                         & \textbf{contrastive}                                               & \textbf{36\%}                                                 \\
\rowcolor[HTML]{FFFFFF} 
top                                                             & efficientNetB0                                                  & triplet                                                            & 31\%                                                          \\
\rowcolor[HTML]{FFFFFF} 
front                                                           & efficientNetB0                                                  & contrastive                                                        & 10\%                                                          \\
\rowcolor[HTML]{FFFFFF} 
front                                                           & efficientNetB0                                                  & triplet                                                            & 30\%                                                          \\ \hline
\end{tabular}}
\end{table}

On the other hand, using VGG, we obtained more relevant results, as seen in Table \ref{tab:vgg}. VGG significantly outperformed EfficientNetB0, and the triplet loss function showed better performance compared to contrastive loss. Additionally, photos from the "top" view yielded superior results.

After recognizing the efficiency of VGG, we conducted additional tests combining images from both the front and top perspectives, referred to as the "all" configuration. However, this approach produced poor results, with accuracies dropping significantly compared to using either front or top views independently. 

This outcome suggests that the distributions of the two perspectives are substantially different, making it challenging for the network to effectively learn from the combined dataset in its current form. These findings highlight the need for further dataset refinement or architectural adjustments to improve performance when utilizing mixed perspectives, guiding future tests and training strategies.

\begin{table}[ht]
\tbl{Preliminary executions using VGG16\label{tab:vgg}}{
\begin{tabular}{llll}
\hline
\textbf{Photo Type} & \textbf{Base Model} & \textbf{Loss Function} & \textbf{Acccuracy} \\ \hline
top                 & VGG16               & contrastive            & \textbf{92\%}      \\
top                 & VGG16               & triplet                & 97\%               \\
front               & VGG16               & contrastive            & 59\%               \\
front               & VGG16               & triplet                & 79\%               \\
all                 & VGG16               & contrastive            & 33\%               \\
all                 & VGG16               & triplet                & 31\%               \\ \hline
\end{tabular}}
\end{table}

\subsection{Final Tests}

With insights from the initial round of tests, we conducted more in-depth experiments increasing the number of animal instances to 26 and utilizing the complete dataset. First, we tested the application of MobileNetV3Large along with VGG16 as base models, optimizing the learning rate for the Adam optimizer. These tests were conducted using the contrastive loss function. Two learning rates, 0.001 and 0.0001, were evaluated. For both MobileNetV3Large and VGG16, 0.0001 proved to be the more efficient choice, yielding higher accuracies. 

However, during these tests, the front-view images began to exhibit poor performance compared to the top-view images. The results of these experiments are summarized in Table \ref{tab:lr1}. In the tables, MobileNetV3Large is referred to simply as MobileNet to facilitate text formatting and improve readability.

\begin{table}[ht]
\tbl{Learning Rate tests with contrastive loss\label{tab:lr1}}{
\begin{tabular}{llll}
\hline
\textbf{Photo Type} & \textbf{Base Model (Loss)}       & \textbf{Learning Rate} & \textbf{F1 Score} \\ \hline
top                 & VGG16 (contrastive)              & 0.001                  & 0.8809            \\
\textbf{top}        & \textbf{VGG16 (contrastive)}     & \textbf{0.0001}        & \textbf{0.9261}   \\
top                 & MobileNet (contrastive)          & 0.001                  & 0.8479            \\
\textbf{top}        & \textbf{MobileNet (contrastive)} & \textbf{0.0001}        & \textbf{0.7345}   \\
front               & VGG16 (contrastive)              & 0.001                  & 0.6938            \\
\textbf{front}      & \textbf{VGG16 (contrastive)}     & \textbf{0.0001}        & \textbf{0.7543}   \\
front               & MobileNet (contrastive)          & 0.001                  & 0.3568            \\
\textbf{front}      & \textbf{MobileNet (contrastive)} & \textbf{0.0001}        & \textbf{0.5651}   \\ \hline
\end{tabular}}
\end{table}

We performed the same learning rate tests using the triplet loss function to evaluate its performance with MobileNetV3Large and VGG16 as base models. The results of these experiments are summarized in Table \ref{tab:lr2}. In these experiments, all results were inferior regardless of the learning rate used, with accuracies significantly lower than those achieved with the contrastive loss function. Consequently, we decided to abandon the triplet loss for further experiments, despite its promising performance in the initial tests. A likely explanation for this decline is that the applied models do not scale well with this type of loss function when the dataset increases in complexity and diversity.

\begin{table}[ht]
\tbl{Learning Rate tests with triplet loss\label{tab:lr2}}{
\begin{tabular}{llll}
\hline
\textbf{Photo Type} & \textbf{Base Model (Loss)}   & \textbf{Learning Rate} & \textbf{F1 Score} \\ \hline
top                 & VGG16 (triplet)              & 0.001                  & 0.3727            \\
\textbf{top}        & \textbf{VGG16 (triplet)}     & \textbf{0.0001}        & \textbf{0.4239}   \\
top                 & MobileNet (triplet)          & 0.001                  & 0.3470            \\
\textbf{top}        & \textbf{MobileNet (triplet)} & \textbf{0.0001}        & \textbf{0.4537}   \\
\textbf{front}      & \textbf{VGG16 (triplet)}     & \textbf{0.001}         & \textbf{0.3436}   \\
front               & VGG16 (triplet)              & 0.0001                 & 0.2799            \\
\textbf{front}      & \textbf{MobileNet (triplet)} & \textbf{0.001}         & \textbf{0.2126}   \\
front               & MobileNet (triplet)          & 0.0001                 & 0.2026            \\ \hline
\end{tabular}}
\end{table}

Using VGG16 and MobileNetV3Large as base models with contrastive loss and learning rate 0.0001, we tested data augmentation to evaluate whether increasing dataset variability could improve F1 scores. Data augmentation introduces diversity to the training data, helping models generalize better by simulating real-world variations. The augmentations applied included horizontal flipping, random rotation between -20 and 20 degrees, and Additive Gaussian Noise with a scale of (0,0.05×255). These augmentations duplicate the dataset size.

These augmentations resulted in a slight improvement in VGG performance. However, while the results for front view images remained insufficient, we achieved a promising F1 score for top view images. Table \ref{tab:lr3} presents a comparison of the best augmentation configuration for the model against the results obtained without augmentation, highlighting the impact of these techniques on overall performance.

\begin{table}[ht]
\tbl{Data Augmentation Tests\label{tab:lr3}}{
\begin{tabular}{llll}
\hline
\textbf{Photo Type} & \textbf{Base Model} & \textbf{Augmentation} & \textbf{F1 Score} \\ \hline
top                 & VGG16                      & none                  & 0.9261            \\
\textbf{top}        & \textbf{VGG16}             & \textbf{rotation}     & \textbf{0.9344}   \\
\textbf{top}        & \textbf{MobileNetV3Large}  & \textbf{none}         & \textbf{0.8479}   \\
top                 & MobileNetV3Large           & flip                  & 0.8360            \\
\textbf{front}        & \textbf{VGG16}             & \textbf{rotation}     & \textbf{0.7724}   \\
front                 & VGG16                      & none                  & 0.7543            \\
\textbf{front}        & \textbf{MobileNetV3Large}  & \textbf{none}         & \textbf{0.5651}   \\
front                 & MobileNetV3Large           & rotation              & \textbf{0.5266}   \\ \hline
\end{tabular}}
\end{table}

In conclusion, our experiments demonstrate that the combination of VGG16 with contrastive loss and top-view images yields the most effective results for the task of cat re-identification. Table \ref{tab:lr4} highlights the top five configurations, showcasing the superior performance achieved with top-view images. This underscores the importance of viewpoint selection and model optimization in maximizing performance. While front-view results remain limited, the outcomes for top-view images confirm the viability of our approach for practical applications

\begin{table}[ht]
\tbl{Top 5 best configurations for our model\label{tab:lr4}}{
\begin{tabular}{llll}
\hline
\textbf{Photo Type} & \textbf{Base Model} & \textbf{Augmentation} & \textbf{F1 Score} \\ \hline
top                 & vgg                        & rotation              & 0.9344            \\
top                 & vgg                        & none                  & 0.9261            \\
top                 & vgg                        & flip                  & 0.9243            \\
top                 & MobileNetV3Large           & False                 & 0.8480            \\
top                 & MobileNetV3Large           & flip                  & 0.8360            \\ \hline
\end{tabular}}
\end{table}

\section{conclusions and future work}
\label{sec:future}
This study made several significant contributions to the field of animal instance re-identification. We developed a new dataset tailored for this purpose, collecting and organizing images of street cats in both top and front perspectives and to support this effort, we implemented a web scraping script capable of efficiently extracting images from the HelloStreetCat livestream. 

Our modular codebase enables seamless testing of various models, loss functions, and augmentation strategies, providing a flexible foundation for future experiments. Together, these contributions establish a comprehensive framework for advancing research in animal re-identification and pave the way for more robust and scalable solutions.

In our experiments, our approach achieved an accuracy of 95.18\%, surpassing the previously reported 94.53\% accuracy by Cho et al. \cite{cho2023multi}, who employed EfficientNetV2 as a feature extractor and SVM as a classifier. It is important to highlight that these results were obtained on different datasets, which means direct comparisons should be interpreted with caution. While our model also uses Siamese networks, as in the method proposed by Li et al. \cite{li2022cat}, which achieved an accuracy of 72.91\%, the higher accuracy observed in our results may reflect differences in dataset characteristics, preprocessing strategies, and overall methodology. These findings suggest that our approach holds promise but require further validation across diverse datasets to fully establish its robustness and generalizability.

For future work, we propose developing an interface capable of automatically analyzing live streams by running the AI model in real-time and saving detected appearances directly into a structured wiki. This would streamline the process of cataloging instances and provide a robust and automated system for tracking animal appearances. 
Additionally, other base models could be trained and evaluated to compare their performance against the current results, potentially uncovering models better suited for the task.

Another promising avenue would be to identify a model better suited for front-view images and develop a mechanism to determine the type of input image—front or top—before processing. This approach would allow the system to select the most appropriate pre-trained model for the given perspective, maximizing accuracy and robustness. By leveraging models specialized for each view, the system could utilize complementary information from both perspectives, leading to more precise and reliable re-identification outcomes.

Furthermore, the dataset can continue to be expanded to include all known instances of cats, ensuring comprehensive coverage for future applications.

By maintaining an updated and diversified dataset, the models could benefit from greater generalization and robustness, improving their real-world applicability. These advancements aim to enhance the system's scalability and reliability while fostering further progress in the field of animal instance re-identification.

\bibliographystyle{ACM-Reference-Format-Journals}
\bibliography{main}


\end{document}